\documentclass[pmlr,tablecaption=bottom]{jmlr}

\usepackage{booktabs}
\usepackage{caption}
\jmlrvolume{124}
\jmlryear{2021}
\jmlrworkshop{NeurIPS 2020 Competition and Demonstration Track}

\title[Solving BBO via Learning Search Space Partition for Local Bayesian Optimization]
{Solving Black-Box Optimization Challenge \\
via Learning Search Space Partition \\
for Local Bayesian Optimization}

\author{
    \Name{Mikita Sazanovich} \Email{mikita.sazanovich@jetbrains.com} \\
    \addr{JetBrains Research} \\
    \addr{HSE University}
    \AND
    \Name{Anastasiya Nikolskaya} \Email{nastia.nikolskaya@gmail.com} \\
    \addr{JetBrains Research} \\
    \addr{Saint Petersburg State University}
    \AND
    \Name{Yury Belousov} \Email{yury-belousov@outlook.com} \\
    \addr{JetBrains Research} \\
    \addr{HSE University}
    \AND
    \Name{Aleksei Shpilman} \Email{alexey@shpilman.com} \\
    \addr{JetBrains Research} \\
    \addr{HSE University}
}

\editors{Hugo Jair Escalante and Katja Hofmann}

\begin{document}

\maketitle

\begin{abstract}
Black-box optimization is one of the vital tasks in machine learning, since it approximates real-world conditions, in that we do not always know all the properties of a given system, up to knowing almost nothing but the results. This paper describes our approach to solving the black-box optimization challenge at NeurIPS 2020 through learning search space partition for local Bayesian optimization. We describe the task of the challenge as well as our algorithm for low budget optimization that we named \texttt{SPBOpt}. We optimize the hyper-parameters of our algorithm for the competition finals using multi-task Bayesian optimization on results from the first two evaluation settings. Our approach has ranked third in the competition finals.
\end{abstract}

\begin{keywords}
black-box optimization challenge, hyper-parameter optimization, Bayesian optimization, learned space partitioning
\end{keywords}

\section{Introduction}

Optimization of hyper-parameters for machine learning models is a common practice. Sometimes, it is done manually, but it could also be automated. Optimizing machine learning models while treating them as a black-box function is a part of black-box optimization. It has been successfully used for many different tasks such as hyper-parameter tuning for convolutional neural networks (\cite{NIPS2012_05311655}), policy optimization in reinforcement learning (\cite{Wang2020LearningSS}), neural architecture search (\cite{Wang2019SampleEfficientNA}).

The black-box optimization challenge (\cite{BBOChallenge}) focuses on applying Bayesian optimization to tuning the hyper-parameters of machine learning models. In this competition, the participants are tasked to optimize the hyper-parameters of an unknown objective function $f$. The algorithm is provided with the hyper-parameter configuration space: the number of hyper-parameters, their types (integer, real, categorical, or boolean), their spaces (linear, logarithmic, logit, or bi-logarithmic), and the lists or the ranges of possible values.  The algorithm should run for $K=16$ iterations and suggest $B=8$ hyper-parameter sets (or points) $x_{k1}, ..., x_{kB}$ per iteration. It receives the value of the objective function for each of them, i.e., $y_{k1}=f(x_{k1}), ..., y_{kB}=f(x_{kB})$. The algorithm is expected to understand how the objective function value depends on the hyper-parameter values from previous iterations' results. The final goal of the algorithm is to minimize the objective function value $f$.

The competition occurs in three evaluation stages. In the first stage, participants can evaluate their algorithms locally by using the \texttt{bayesmark} package. In the second stage, participants can submit a limited number of attempts to the remote evaluation server, which participants could use throughout the competition. One submission from each participant is evaluated on a hidden set of test cases in the final stage.

The competition uses both realistic and synthetic machine learning problems as test cases. By combining it with the low-budget setting, i.e., the low number of optimization iterations, the challenge serves as a testing ground for algorithms that could be applied to real-world model tuning.

This paper describes our \texttt{SPBOpt} algorithm that tackles the challenge. It uses both the local and remote evaluation settings to tune itself for the competition finals. We also provide a detailed description of the training process. Our algorithm has ranked third in the finals on the hidden set of test cases.

\section{Related Work}

In our solution, we follow the Bayesian Optimization (BO) approach. BO approaches allow to fit black-box objective functions in a derivative-free way. The algorithms are typically based on the surrogate models and utilize the so-called acquisition functions. In the classic \cite{NIPS2012_05311655} paper, the authors apply BO to perform hyper-parameters tuning of general machine learning algorithms. They use a Gaussian process as a surrogate model and consider an acquisition function based on the Expected Improvement criterion.

One issue of the BO approaches is that they usually work well only when provided with sufficient data while the competition is held in the low-budget environment. Many algorithms have been proposed as an improvement over the base BO to tackle this problem.

One direction is to replace the Gaussian process surrogate model. The authors of \cite{NIPS2011_86e8f7ab}, anticipating that hyper-parameter tuning of machine learning models will lead to a high-dimensional problem with a small optimization budget, propose to use the tree-structured Parzen estimator (TPE). This approach has now been integrated into popular BBO libraries such as \texttt{hyperopt} (\cite{pmlr-v28-bergstra13}). In \cite{hutter2011} the authors use random forests, \cite{snoek2015scalable} proposes to use Bayesian linear regression on features from neural networks. In this paper, we present a comparison with the \texttt{hyperopt} in the local evaluation setting.

Another direction of improving the performance in the low-budget setting is the restriction of the optimization region. The authors of \cite{Eriksson2019Turbo} propose to use a local probabilistic approach for global optimization of large-scale high-dimensional problems. We find that, although the end goal of our solution is different, the local Bayesian modeling works well for the low-budget setting due to its sample efficiency. Constraining the optimization region could also be done by learning where to search for the optimal parameters. \cite{Wang2020LearningSS} propose to learn the partition of the search space. We follow this direction in our algorithm.

\section{Algorithm}

\begin{figure}[htbp]
  \centering
  \includegraphics[width=0.4\textwidth]{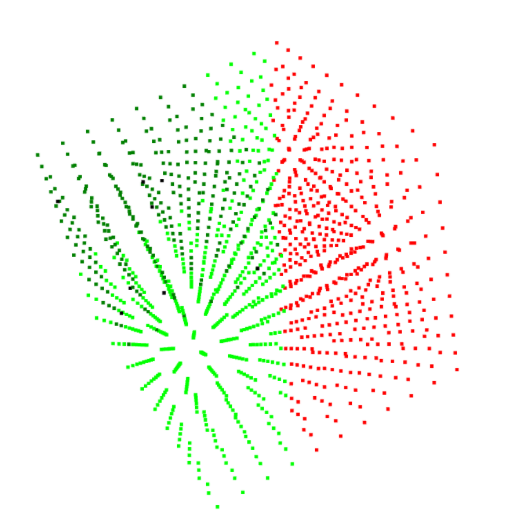}
  \caption{An example space partition for a 3D hyper-parameter space. The red points lie outside the selected region from the partition. The light green points are inside it but outside the region of the local Bayesian optimization model. The dark green points are some of the points which could be suggested next. In the end, the algorithm selects the black points for the next iteration.}
  \label{img-space-partitioning}
\end{figure}

\subsection{Overview}

To tackle this challenge, we have developed the \texttt{SPBOpt} algorithm. It consists of several parts, including the initial sampling method, local Bayesian optimization, and learning search space partition.

First, we run the initial points generator to provide $n_{init}$ points for our model to start. The $n_{init}$ value could depend on the number of hyper-parameters or be a predetermined number. The initialization usually runs from 1 to 4 iterations. After the initialization, the algorithm builds the space partitioning. The space partitioning is rebuilt every $n_{rebuild}$ iterations after its first construction. When the space partitioning is built or rebuilt, the local Bayesian optimization model runs in the space region with the lowest average objective function value. It is initialized from all previously evaluated points in the region, and it runs for $n_{rebuild}$ iterations. We set $n_{rebuild}$ to 4.

Furthermore, we reset the algorithm to its initial state (i.e., remove all accumulated points, its space partitioning, and begin from the initialization) every $n_{reset}$ iterations if no progress has been made. To measure the progress, we consider the minimum function objective before and after the last $n_{reset}$ iterations. If the minimum value is the same, we say that no progress has been made and reset our algorithm. It helps with getting out of the local minimum and making progress in the global optimization task. We set $n_{reset}$ to 8.

\subsection{Initial sampling method}

One of the approaches for initial point generation is to generate them completely randomly. The downside of this is that there is no guarantee that the points are spread well enough across all the dimensions. Sampling methods such as Latin hypercube, Sobol, Halton, Hammersly (\cite{Greenhill2020BayesianOF}) and MaxPro (\cite{Joseph2015MaximumPD}) take advantage of the fact that we know beforehand how many random points we want to sample. Then the initial points can be sampled in a way that explores each dimension. We experiment with these methods in our algorithm. Our final submission uses a variant of Latin hypercube sampling.

\subsection{Local Bayesian optimization}

\label{sub-local-bayesian-model}

We use the trust-region Bayesian optimization (\texttt{TuRBO}) algorithm from \cite{Eriksson2019Turbo} as our local Bayesian optimization model. Considering the low budget for the number of iterations, we modify it with a decay factor, which shrinks the trust region. We use the policy to decay the region side lengths by a constant factor $decay$ with each iteration if we have already used half of our iterations budget, i.e., past eight iterations.

\subsection{Learning search space partition}

We intend on learning the space partition into regions with high/low objective function values, similar to \cite{Wang2020LearningSS}. Using the space partition, we select the region with the lowest average objective function value and run the local Bayesian optimization algorithm described above. Fig. \ref{img-space-partitioning} shows an example of such a space partition during a run of the algorithm.

More formally, when we construct a space partitioning at an iteration $t$ (from 1 to $K$), we have a dataset $D_t$ which consists of previously evaluated points $(x_1, y_1)$, ..., $(x_{n_t}, y_{n_t})$, where $n_t=t*B$. We recursively split the current set of points into a left and a right sub-tree. The split is built as follows:
\begin{enumerate}
    \item We run the KMeans algorithm to group the points into 2 clusters based on their objective function values $y_i$. The left sub-tree is formed from a set of points with a lower average function objective value.
    \item Using KMeans algorithm labels as ground-truth labels, we train a split model to predict whether a set of hyper-parameters would fall into the first or the second cluster. We consider SVM with different kernels and k-nearest neighbors algorithm for the split model.
    \item The split model filters the current set of points so that only the points which are predicted to be in the left sub-tree remain.
\end{enumerate}

We continue to split the set of points until we reach the maximum depth $max_{depth}=5$, or the new set of points is not large enough for the initialization of the local Bayesian optimization model.

\subsection{Optimization}

\label{sub-optimization}

To find the best hyper-parameters of our algorithm for the final evaluation, we use the multi-task Bayesian optimization method from \cite{JMLR:v20:18-225} to build a multi-task Gaussian process to combine the evaluation results in local and remote settings. We provide the results of the optimization in Section \ref{subsec-candidates-selection}.

\section{Experiments}

\begin{figure}
    \centering
    \includegraphics[width=\textwidth]{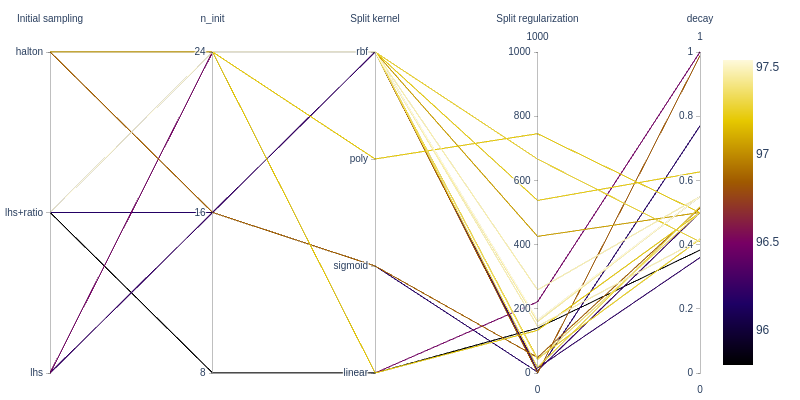}
    \caption{Parallel coordinates diagram for the \texttt{SPBOpt} hyper-parameters and remote score. lhs stands for Latin hypercube sampling, lhs+ratio further optimizes it using the ratio criterion.}
    \label{fig:parallel_diagram}
\end{figure}

\subsection{Scoring}

The challenge uses the following scoring protocol. The score given to a run of an algorithm on an objective function is the minimum value received by an algorithm normalized by the expected minimum and maximum objective function values. The score is then multiplied by 100. More formally, let $f_{a}$ be a minimum value received by an algorithm, $f_{min}$ be the expected minimum  and $f_{max}$ the expected maximum values. The score is computed as follows: $s_{a}=100*(1.0-\frac{f_{a}-f_{min}}{f_{max}-f_{min}})=100*\frac{f_{max}-f_{a}}{f_{max}-f_{min}}$.

\subsection{Evaluation settings}

The scoring of the algorithm happens in three different settings.

The local evaluation is done by the \texttt{bayesmark} package. The evaluation consists of 108 experiments. Each experiment includes tuning the hyper-parameters of machine learning models such as decision trees, random forests, SVM, k-nearest neighbors for regression or classification tasks on several datasets. The result of the evaluation is the average of the scores received by the algorithm.

The details of the remote evaluation setting are hidden from the participants. After the competition, the organizers disclosed that they had 60 experiments for tuning hyper-parameters of machine learning models. The results of the evaluation are the average of the scores for the experiments.

In the finals, the participants were evaluated on a hidden set of 60 different experiments.

Overall, the participants are expected to test their algorithms extensively using the local evaluation, validate their solutions in the remote evaluation setting, and select a single algorithm for the finals. The goal is to create an algorithm that will perform well in all settings and will be resilient to changes in the experimental setup.

\subsection{Candidates selection}

\label{subsec-candidates-selection}

We have run a preliminary analysis of various hyper-parameter settings for our algorithm in both local and remote evaluation settings. From the remote evaluation results, we have built the diagram which depicts the dependence between hyper-parameters and the remote score. The diagram is shown in Figure \ref{fig:parallel_diagram}. We can see how changing the internal configuration of our algorithm could change the remote score from 95.798 to 97.537. We run a similar local evaluation.

During the analysis, we found that the variants of the Latin hypercube sampling, higher number of initialization points, polynomial and radial basis function SVM kernels work better with our algorithm.

To account for both settings and not overfit specifically to the remote setting, we build the multi-task model as described in Section \ref{sub-optimization}. Then we generate 5 candidates for the pool of finals candidates. Table \ref{tab-hyper-parameters} shows the hyper-parameters of the generated set of candidates. The next section presents the candidates' evaluation results and our selection process for the competition finals.

\begin{table}[htbp]
  \small
  \caption{Hyper-parameters of the finals candidates.}
  \label{tab-hyper-parameters}
  \centering
  \begin{tabular}{ccccccc}
    \toprule
    Configuration & Initial sampling & $n_{init}$ & Split model & Split kernel & Split regularization & $decay$ \\
    \midrule
    \texttt{SPBOpt}\textsubscript{1} & lhs+ratio & 8 & SVM & rbf & 0.002762 & 0.700 \\
    \texttt{SPBOpt}\textsubscript{2} & lhs+ratio & 24 & SVM & poly & 745.322745 & 0.499 \\
    \texttt{SPBOpt}\textsubscript{3} & lhs+ratio & 24 & SVM & rbf & 145.415497 & 0.416 \\
    \texttt{SPBOpt}\textsubscript{4} & lhs+ratio & 24 & SVM & rbf & 165.066908 & 0.549 \\
    \texttt{SPBOpt}\textsubscript{5} & lhs+ratio & 24 & SVM & rbf & 76.7041709 & 0.677 \\
    \bottomrule
  \end{tabular}
\end{table}

\subsection{Local and remote evaluation of candidates}

We evaluate the set of candidates from Section \ref{subsec-candidates-selection} and the competition baselines in the local evaluation setting. The competition baselines include optimization algorithms, namely: random search, \texttt{nevergrad}, \texttt{opentuner}, \texttt{hyperopt} (\cite{pmlr-v28-bergstra13}), \texttt{skopt}, \texttt{TuRBO} (\cite{Eriksson2019Turbo}) and \texttt{pysot} (\cite{Eriksson2019pySOTAP}). The methods were selected as the state-of-the-art out-of-the-box optimization approaches. We run each baseline and our configurations locally 8 times.

Table \ref{tab-local-evaluation} shows the local results. Note that all configurations of \texttt{SPBOpt} outperform the competition baselines. In particular, our approach of learning search space partition for local Bayesian optimization performs better than the \texttt{TuRBO} algorithm.

Using the remote evaluation setting for validation, we send the same set of candidates from Section \ref{subsec-candidates-selection} to a remote evaluation server. We evaluate each candidate remotely 3 times.

Table \ref{tab-remote-evaluation} presents the remote results. Most of the \texttt{SPBOpt} configurations score more than 97 points on average.

\begin{table}[htbp]
  \small
  \caption{Local evaluation of the methods.}
  \label{tab-local-evaluation}
  \centering
  \begin{tabular}{ccc}
    \toprule
    Method & Local scores mean & Local scores stddev  \\
    \midrule
    random search & 91.658 & 0.572 \\
    \texttt{nevergrad} & 92.765 & 0.912 \\
    \texttt{opentuner} & 93.693 & 0.711 \\
    \texttt{hyperopt} & 95.881 & 0.429 \\
    \texttt{skopt} & 96.670 & 0.528 \\
    \texttt{TuRBO} & 97.765 & 0.437 \\
    \texttt{pysot} & 98.200 & 0.480 \\
    \midrule
    \texttt{SPBOpt}\textsubscript{1} & 98.306 & 0.550 \\
    \texttt{SPBOpt}\textsubscript{2} & \textbf{98.898} & 0.316 \\
    \texttt{SPBOpt}\textsubscript{3} & 98.555 & 0.474 \\
    \texttt{SPBOpt}\textsubscript{4} & 98.738 & 0.661 \\
    \texttt{SPBOpt}\textsubscript{5} & 98.538 & 0.308 \\
    \bottomrule
  \end{tabular}
\end{table}

\begin{table}[htbp]
  \small
  \centering
  \caption{Remote evaluation of the finals candidates.}
  \label{tab-remote-evaluation}
  \begin{tabular}{c c c}
    \toprule
    Configuration & Remote scores mean & Remote scores stddev \\
    \midrule
    \texttt{SPBOpt}\textsubscript{1} & 96.939 & 0.300 \\
    \texttt{SPBOpt}\textsubscript{2} & \textbf{97.557} & 0.281 \\
    \texttt{SPBOpt}\textsubscript{3} & 97.451 & 0.257 \\
    \texttt{SPBOpt}\textsubscript{4} & 97.345 & 0.167 \\
    \texttt{SPBOpt}\textsubscript{5} & 97.505 & 0.117 \\
    \bottomrule
  \end{tabular}
\end{table}

\subsection{Competition finals}

We have selected Candidate 2 for the competition finals based on the evaluation scores of all runs from local and remote evaluations using the Wilcoxon signed-rank test with \emph{p}-value less than $0.05$.

The results of the top competitors are shown in Table \ref{tab-finals-results}. Our approach has scored 92.509 in the finals and has placed third overall. All participants have scored fewer points in the final round when compared to the remote evaluation. This fact suggests, that most algorithms overfit to the remote setting, and more resilient algorithms, including our \texttt{SPBOpt}, have proved their prominence in the finals.

The organizers have also shared the finals results for the random search and the best performing baseline, \texttt{TuRBO}. We can see that \texttt{SPBOpt} outperforms \texttt{TuRBO} in this setting too.

\begin{table}[htbp]
  \small
  \centering
  \caption{Results of the competition finals.}
  \label{tab-finals-results}
  \begin{tabular}{c c}
    \toprule
    Method & Score \\
    \midrule
    \cite{BBO1} & 93.519 \\
    \cite{BBO2} & 92.928 \\
    \texttt{SPBOpt} & 92.509  \\
    \cite{BBO4} & 92.212 \\
    \cite{BBO5} & 91.806 \\
    \midrule
    \texttt{TuRBO} & 88.921 \\
    random search & 75.404 \\
    \bottomrule
  \end{tabular}
\end{table}

\section{Conclusion}

In this paper, we present our approach to the black-box optimization challenge, the \texttt{SPBOpt} algorithm. It includes local Bayesian optimization and learning search space partition. We optimize the hyper-parameters of our algorithm by using another black-box optimization algorithm. The \texttt{SPBOpt} algorithm has ranked third in the Black-Box Optimization competition finals at NeurIPS 2020. We also demonstrate how the configuration of our algorithm impacts the final result and provide reasonable default choices.

We believe that our approach could be applied to tuning other real-world models due to its low overhead and focus on sample efficiency.

The code for our solution is available at:
\\
\url{https://github.com/jbr-ai-labs/bbo-challenge-jetbrains-research}.

\acks{We would like to acknowledge support for this project
from JetBrains Research.}

\bibliography{common}

\end{document}